\documentclass{Interspeech2024}
\interspeechcameraready 

\title{MultiTalk: Enhancing 3D Talking Head Generation \\Across Languages with Multilingual Video Dataset}

\name[affiliation={1*}]{Kim}{Sung-Bin}{}
\name[affiliation={2*}]{Lee}{Chae-Yeon}{}
\name[affiliation={1*}]{Gihun}{Son}{}
\name[affiliation={1}]{Oh}{Hyun-Bin}{newline}
\name[affiliation={3}]{Janghoon}{Ju}{}
\name[affiliation={3}]{Suekyeong}{Nam}{}
\name[affiliation={1,2,4}]{Tae-Hyun}{Oh}{}

\address{
  $^1$Dept.~of Electrical Engineering and $^{2}$Grad.~School of Artificial Intelligence, POSTECH, Korea\\
  $^3$KRAFTON, Korea \\
  $^4$Institute for Convergence Research and Education in Advanced Technology, Yonsei University, Korea
}
  
\email{\{sungbin, chaeyeon.lee, gihun.son, taehyun\}@postech.ac.kr}

\keywords{Speech-driven 3D talking head, Video dataset, Multilingual, Audio-visual speech recognition}

\usepackage{hyperref}
\usepackage{enumitem}
\usepackage{multirow}
\usepackage{wrapfig}
\usepackage{colortbl}
\usepackage[normalem]{ulem}
\usepackage{microtype}
\usepackage{comment}
\usepackage[symbol]{footmisc}

\setlist[itemize]{align=parleft,left=0pt,topsep=1mm,itemsep=0mm,parsep=1mm}

\definecolor{azure(colorwheel)}{rgb}{0.0, 0.5, 1.0}
\definecolor{nicegreen}{rgb}{0.0, 0.7, 0.1}
\definecolor{yw}{rgb}{0.01176, 0.5490, 0.5490}
\definecolor{ashblue}{rgb}{0.36, 0.54, 0.66}
\definecolor{ashgrey}{rgb}{0.7, 0.75, 0.71}
\definecolor{applegreen}{rgb}{0.55, 0.71, 0.0}
\definecolor{blue}{rgb}{0.0, 0.0, 1.0}
\definecolor{postechred}{rgb}{0.784, 0.003, 0.313}
\definecolor{ywg}{rgb}{0.9960, 0.8984, 0.5859}
\definecolor{ballblue}{rgb}{0.13, 0.67, 0.8}
\definecolor{cornellred}{rgb}{0.7, 0.11, 0.11}
\definecolor{darkcyan}{rgb}{0.0, 0.55, 0.55}
\definecolor{CuGray}{gray}{0.9}
\definecolor{airforceblue}{rgb}{0.36, 0.54, 0.66}
\definecolor{rev}{rgb}{0.784, 0.003, 0.313}
\definecolor{pink}{cmyk}{0, 0.7808, 0.4429, 0.1412}
\definecolor{amethyst}{rgb}{0.6, 0.4, 0.8}
\definecolor{black}{rgb}{0.0, 0.0, 0.0}
\definecolor{tb3_yellow}{rgb}{0.996, 1.0, 0.6}
\definecolor{tb3_orange}{rgb}{0.980, 0.8, 0.604}
\definecolor{tb3_red}{rgb}{0.972, 0.6, 0.6}
\definecolor{dimgray}{rgb}{0.41, 0.41, 0.41}
\definecolor{brickred}{rgb}{0.8, 0.25, 0.33}
\definecolor{bleudefrance}{rgb}{0.19, 0.55, 0.91}
\definecolor{blue(ncs)}{rgb}{0.265, 0.445, 0.765}
\definecolor{blue(ryb)}{rgb}{0.01, 0.28, 1.0}
\definecolor{orange}{rgb}{1.0, 0.49, 0.0}
\definecolor{Gray}{gray}{0.88}
\definecolor{green(ncs)}{rgb}{0.0, 0.62, 0.42}
\definecolor{brightpink}{rgb}{1.0, 0.0, 0.5}

\definecolor{kellygreen}{rgb}{0.3, 0.73, 0.09}

\newcolumntype{g}{>{\columncolor{CuGray}}c}
\newcolumntype{z}{>{\columncolor{CuGray}}l}

\renewcommand{\paragraph}[1]{\vspace{1mm}\noindent\textbf{#1.}\,\,}

\usepackage{xspace}

\makeatletter
\def\@fnsymbol#1{\ensuremath{\ifcase#1\or *\or \dagger\or \ddagger\or
   \mathsection\or \mathparagraph\or \|\or **\or \dagger\dagger
   \or \ddagger\ddagger \else\@ctrerr\fi}}
\makeatother

\def\onedot{.\@\xspace}
\def\eg{\emph{e.g}\onedot} 
\def\ie{\emph{i.e}\onedot}

\def\etal{\emph{et al}\onedot}

\newcommand{\Sref}[1]{Sec.~\ref{#1}}
\newcommand{\Eref}[1]{Eq.~(\ref{#1})}
\newcommand{\Fref}[1]{Fig.~\ref{#1}}
\newcommand{\Tref}[1]{Table~\ref{#1}}

\newcommand{\calL}{{\mathcal{L}}}

\newcommand{\be}{\begin{eqnarray}}
\newcommand{\ee}{\end{eqnarray}}
\newcommand{\bee}{\begin{eqnarray*}}
\newcommand{\eee}{\end{eqnarray*}}

\newcommand{\matrixb}{\left[ \begin{array}}
\newcommand{\matrixe}{\end{array} \right]}

\newcommand{\argmin}{\operatornamewithlimits{\arg \min}}

\usepackage{mathtools}

\DeclarePairedDelimiter{\norm}{\lVert}{\rVert}

\begin{document}

\maketitle

\begin{abstract}
    Recent studies in speech-driven 3D talking head generation have achieved convincing results in verbal articulations. However, generating accurate lip-syncs degrades when applied to input speech in other languages, possibly due to the lack of datasets covering a broad spectrum of facial movements across languages. In this work, we introduce a novel task to generate 3D talking heads from speeches of diverse languages. We collect a new multilingual 2D video dataset comprising over 420 hours of talking videos in 20 languages. With our proposed dataset, we present a multilingually enhanced model that incorporates language-specific style embeddings, enabling it to capture the unique mouth movements associated with each language. Additionally, we present a metric for assessing lip-sync accuracy in multilingual settings. We demonstrate that training a 3D talking head model with our proposed dataset significantly enhances its multilingual performance. Codes and datasets are available at \url{https://multi-talk.github.io/}.
\end{abstract}

\def\thefootnote{*}\footnotetext{These authors contributed equally.}
\def\thefootnote{\arabic{footnote}}
\section{Introduction}
Speech-driven 3D talking heads are key components in virtual avatars, enhancing realism and improving user engagement in diverse multimedia applications~\cite{liu2009analysis, education, wohlgenannt2020virtual}. Recent advancements in deep learning have significantly advanced the field of 3D talking heads. 
Earlier efforts~\cite{voca, karras2017audio, meshtalk, faceformer, codetalker} have focused on enhancing lip synchronization, while more recent studies aim to enable the expression of various emotions~\cite{emotalk, emote} and non-verbal signals~\cite{laughtalk}, or even to develop personalized models~\cite{imitator}.

However, the multilingual capabilities of 3D talking heads have received less attention and remain underexplored.
Despite claims from previous studies~\cite{faceformer, voca, huang2021speaker} that their models are language-agnostic, we observe that Huang~\etal~\cite{huang2021speaker} cover only two languages, and the quality of the generated meshes
from prior works~\cite{faceformer, voca} 
degrades when input speech deviates from the English language family.
We hypothesize that this limitation stems from the scarcity of diverse 3D talking head datasets. 
Existing datasets, such as VOCASET~\cite{voca} and BIWI~\cite{biwi}, are not only small in scale but also limited in expressiveness, diversity, and language scope (English-only). 
Even a more sophisticated model, designed to handle diverse languages, may be constrained by the styles and motion characteristics of available datasets.

\begin{figure}[t]
    \centering
    \includegraphics[width=1\linewidth]{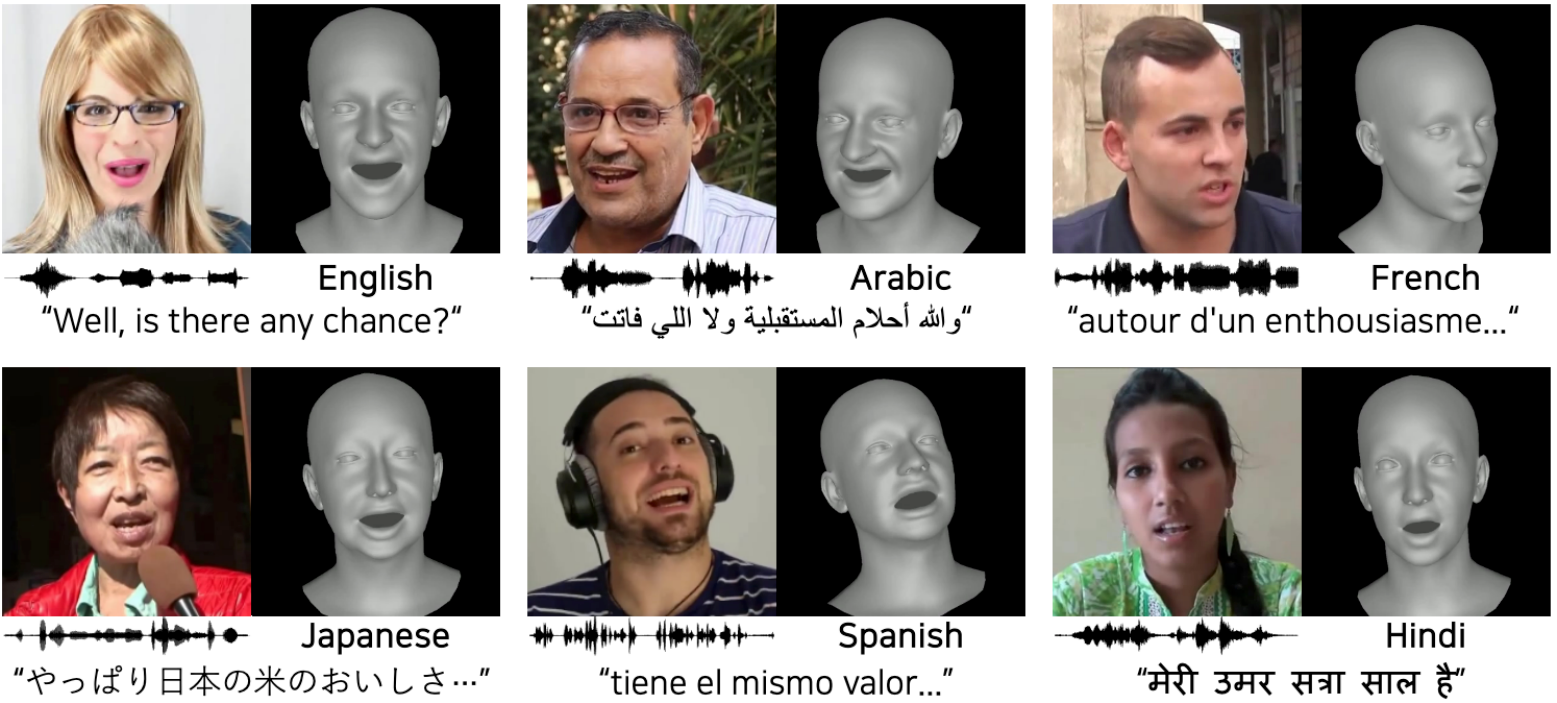}\vspace{-2mm}
    \caption{\textbf{Samples of our MultiTalk dataset.} Each 2D video is annotated with the language type and the pseudo transcript, and a subset of the videos further provides pseudo 3D mesh vertices.}\vspace{-4.4mm}
    \label{fig:data_samples}
\end{figure}

To tackle this challenge, we introduce a novel task of generating 3D talking heads from speeches in diverse languages, \ie, multilingual 3D talking heads. 
For this task, we collect the \textbf{MultiTalk dataset}, comprising in-the-wild 2D talking videos across 20 different languages, paired with corresponding pseudo 3D meshes and transcripts (see \Fref{fig:data_samples}). 
We design an automated data collection pipeline to parse short utterances of frontal talking videos in diverse languages from YouTube. 
As these 2D videos lack 3D metadata, we leverage an off-the-shelf 3D reconstruction model~\cite{spectre} to generate reliable and robust pseudo ground-truth 3D mesh vertices~\cite{flame} for the collected 2D videos.

To demonstrate the effectiveness of our dataset for multilingual 3D talking heads, we introduce a strong baseline model, \textbf{MultiTalk}, by training on a subset of our dataset. 
Inspired by previous works~\cite{codetalker, learning2listen}, we start by training a vector-quantized autoencoder (VQ-VAE)~\cite{vqvae} to learn a discrete codebook, which encodes expressive 3D facial motions across various languages. 
By utilizing this discrete codebook, we then train a temporal autoregressive model to synthesize sequences of 3D faces, conditioned on both the input speech and the learnable language embedding. This language embedding captures the stylistic nuances of facial motions specific to each language family.

We validate our baseline model against existing 3D talking head models~\cite{voca, faceformer, codetalker, selftalk} trained on the English-only VOCASET dataset. 
As this task is novel, we propose a new evaluation metric, Audio-Visual Lip Readability (AVLR), which assesses the lip-sync accuracy of 3D talking heads on multilingual speeches using a pre-trained Audio-Visual Speech Recognition (AVSR) model~\cite{muavic}. 
Through the experiments, we show that our model performs favorably across diverse languages compared to previous works. 
Our main contributions are summarized as follows:
\begin{itemize}
    \item Proposing a new task, multilingual 3D talking head, accompanied by an evaluation metric to measure the lip synchronization accuracy on multilingual speech.
    \item Collecting the MultiTalk dataset, featuring over 420 hours of 2D videos with paired 3D metadata in 20 different languages.
    \item Introducing a strong baseline, MultiTalk, capable of generating accurate and expressive 3D faces from multilingual speech.
\end{itemize}

\section{Learning multilingual 3D talking head}
In this section, we introduce our new multilingual video dataset in \Sref{subsec:dataset} and describe the proposed baseline model for multilingual 3D talking head generation in \Sref{subsec:model}.

\begin{table}[tp]
\caption{\textbf{Statistics of our MultiTalk dataset.} We present a 2D talking video dataset that is well-balanced across 20 languages (each accounting for 2.0-9.7\%), accompanied by pseudo 3D mesh vertices and transcripts for each video.}
\footnotesize
\centering

    \resizebox{0.50\linewidth}{!}{
    \begin{tabular}{c}
    \includegraphics[width=1\linewidth]{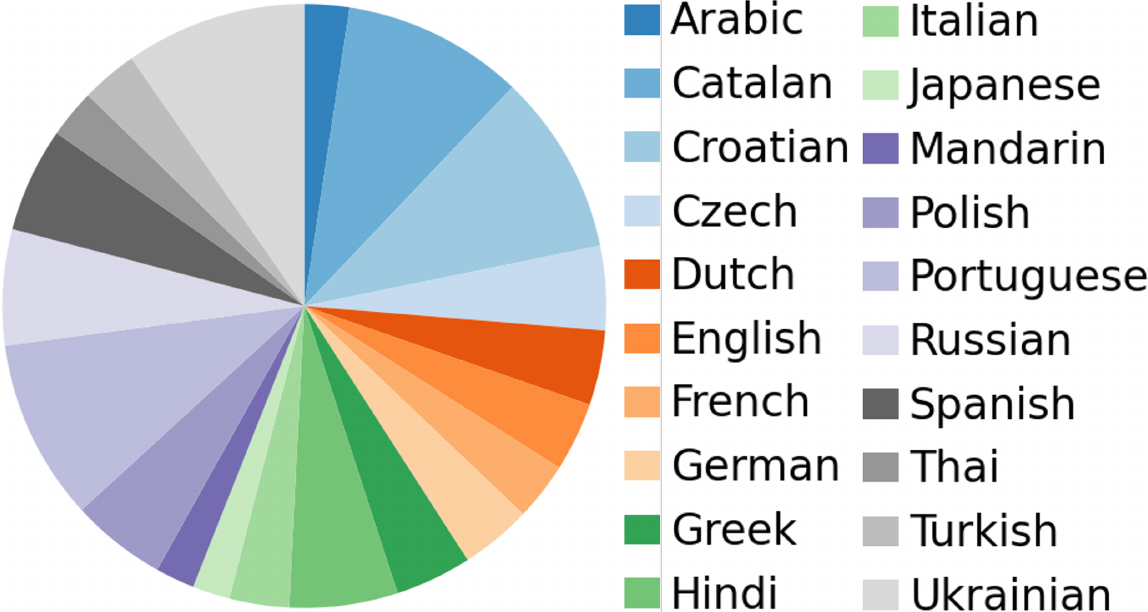}
    \end{tabular}}
    \,
    \resizebox{0.46\linewidth}{!}{
    \begin{tabular}{ll @{\quad}cc}
    \toprule
    \multicolumn{2}{c}{Statistics of our MultiTalk dataset}\\
    \cmidrule{1-2}
    Num. of languages&20\\
    Num. of video clips&294k\\
    Total hours&423.2 h\\
    Avg. duration&5.2 sec.\\
    \bottomrule
    \end{tabular}
    }
    
\label{tab:statistic}
\end{table}

\subsection{MultiTalk dataset} \label{subsec:dataset}
We introduce the MultiTalk dataset, featuring over 420 hours of multilingual 2D talking videos across 20 languages. 
Despite the abundance of 2D video datasets~\cite{yu2023celebv,Chung18b, Nagrani17}, we aim to curate a dataset with more balanced statistics across a broader range of languages.
Each video in the MultiTalk dataset is annotated with the language type of a speech and a pseudo-transcript generated using Whisper~\cite{whisper}, and a subset among the videos is annotated with pseudo 3D mesh vertices. 
Samples and statistics of the MultiTalk dataset are shown in \Fref{fig:data_samples} and \Tref{tab:statistic}, respectively. 
We design an automated pipeline to obtain short utterances of talking videos in diverse languages, described as follows.

\paragraph{Collecting 2D videos}
We begin by designing various queries that incorporate keywords, such as ``nationality,'' ``interviews,'' and ``conversation.'' 
These queries are prompted to YouTube to retrieve in-the-wild human talking 2D videos in different languages and diverse scenarios.

\paragraph{Active speaker verification}
The goal here is to trim a raw video into segments that only contain talking faces synchronized to speech, while removing clips with non-active speakers. 
To achieve this, we leverage TalkNet~\cite{tao2021someone}, which performs audio-visual cross-attention to identify the visible person in speaking. 
We set conservative thresholds to minimize false positives and ensure that only the cleaned video is left. 
We further trim the video when gaps occur during speaking, resulting in short utterances. 

\paragraph{Frontal face verification}
The faces in the filtered videos do not always face the front, which can prevent the model from learning clear facial movements.
Thus, we measure the angle of yaw and pitch of the face using Mediapipe~\cite{mediapipe} and filter out videos with abrupt angle changes (indicating abrupt head movements) or large yaw or pitch angles (side faces).
This process concludes the automated pipeline for collecting cleaned 2D frontal talking face videos with short utterances in diverse languages. We further leverage Whisper~\cite{whisper}\footnote{Referring to \url{https://github.com/openai/whisper}, Whisper's performance on our target language yields a word error rate (WER) of 2.8\% to 17.0\%, which is quite accurate.} trained on each language, to annotate the pseudo-transcript for each video clip.

\paragraph{Lifting 2D video to 3D}
From the subset of collected 2D talking videos, we reconstruct 3D meshes that are synchronized with both the audio and facial movements of the video clips. 
Similar to prior arts~\cite{emote, laughtalk} that demonstrate the effectiveness of pseudo-3D reconstructions for training 3D talking heads, we leverage SPECTRE~\cite{spectre} to reconstruct accurate and robust pseudo 3D meshes from 2D talking videos.
Unlike existing datasets, \eg, VOCASET~\cite{voca}, which are limited in small scale and English-only speeches, our newly annotated 3D mesh dataset encompasses expressive facial motions, paired with speeches that vary in diverse tones and pitches across a wide range of languages.

\begin{figure}[tp]
    \centering
    \includegraphics[width=01\linewidth]{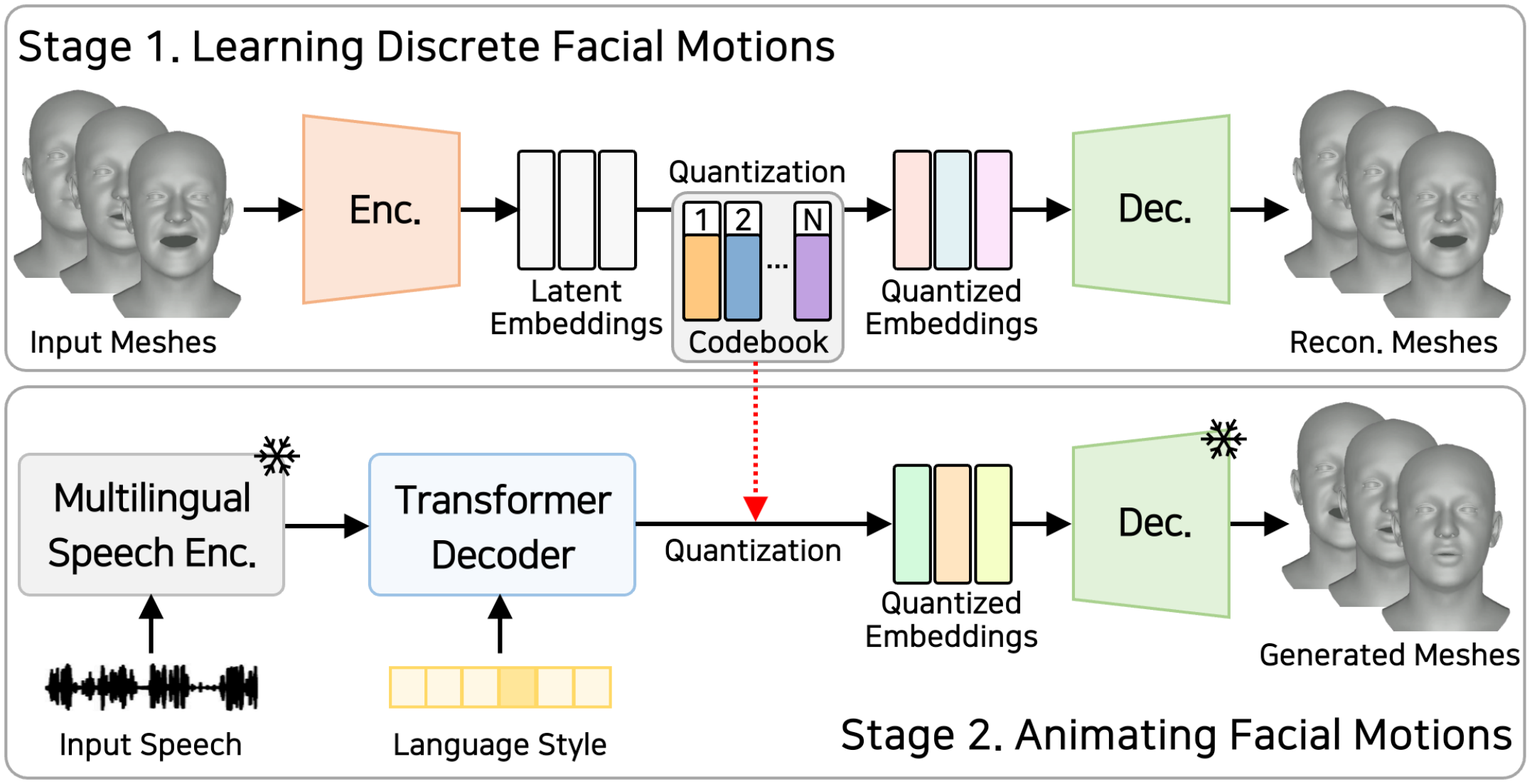}
    \caption{\textbf{Overall pipeline of MultiTalk.} In stage 1, a codebook of discrete motions is learned from 3D facial meshes speaking in diverse languages. In stage 2, the model learns to autoregressively generate a sequence of motion representations
    from an input speech. These representations 
    are quantized by the codebook, thereby synthesizing speech-driven 3D talking head. 
    }
    \label{fig:pipeline}
\end{figure}

\subsection{Speech-driven multilingual 3D talking head}\label{subsec:model}
Using the subset of the dataset we collected, as detailed in \Sref{subsec:dataset}, we aim to develop a model capable of generating accurate 3D talking head synchronized with input speech in various languages.
Despite the increased diversity of the dataset, na\"ively dumping all data into the model could result in learning only average facial movements.
To address this, we break down the task into sub-problems and introduce a baseline model, MultiTalk.
MultiTalk undergoes a two-stage training process: first, learning a general facial motion prior through discrete codes, then training a speech-driven temporal autoregressive model to animate 3D faces with the learned discrete codes.
Echoing the success of previous works~\cite{vqvae, codetalker, learning2listen}, the disentanglement of the learning process allows the model to effectively construct rich discrete motion prior from the diverse talking faces of various languages, which is then leveraged in the second stage for synthesis.

The task formulation is specified as follows: Let $\mathbf{V}_{1:T} = (\mathbf{v}_1, \dots, \mathbf{v}_T)$ denote a temporal sequence of ground-truth facial motions, where each frame $\mathbf{v}_t \in \mathbb{R}^{N \times 3}$ consisting $N$ vertices, represents the 3D facial movement. 
Additionally, let $\mathbf{S}_{1:T} = (\mathbf{s}_1, \dots, \mathbf{s}_T)$ be a sequence of speech representations. 
By conditioning on input speech $\mathbf{S}_{1:T}$, the goal in this task is to sequentially predict facial movements $\mathbf{\hat{V}}_{1:T}$, similar to $\mathbf{V}_{1:T}$.

\paragraph{Learning discrete facial motion}
Following CodeTalker~\cite{codetalker}, we extend the use of vector quantized autoencoder (VQ-VAE)~\cite{vqvae} to learn a discrete codebook of context-rich facial motions
(refer to Stage 1 in \Fref{fig:pipeline}). 
As speech data is not required for training VQ-VAE, we utilize a large amount of 3D motion sequences from the MultiTalk dataset for learning the prior. 
This enables the learned prior to cover a broad spectrum of facial motions observed in speakers of diverse languages.

VQ-VAE consists of an encoder ($E_v$), a decoder ($D_v$), and a discrete codebook $\mathcal{Z}=\{\mathbf{z}_k\}_{k=1}^{K}$, where $\mathbf{z}_k\in\mathbb{R}^{d_z}$, and is trained to self-reconstruct realistic facial motions. 
Specifically, given facial motion sequences in continuous domain $\mathbf{V}_{1:T}$, the VQ-VAE encoder ($E_v$), which is designed with a Transformer~\cite{transformer} layer, first encodes the continuous motion sequences into latent features $\mathbf{\hat{Z}} \in \mathbb{R}^{T'\times d_z}$, where $T'$ denotes the number of frames of downsampled features. 
Subsequently, $\mathbf{\hat{Z}}$ is quantized to $\mathbf{Z^q}$ through an element-wise quantization function $Q_v$ that maps each element in $\mathbf{\hat{Z}}$ to its nearest codebook entry:
\begin{equation}\label{eq:quantize}
\mathbf{Z^q}=Q_v(\mathbf{\hat{Z}})\coloneqq(\argmin_{\mathbf{z}_k\in\mathcal{Z}}\norm{\mathbf{\hat{z}}_{t}-\mathbf{z}_k}_2)\in\mathbb{R}^{T'\times d_z}.
\end{equation}
$\mathbf{Z_q}$ is then reconstructed back into continuous motions $\mathbf{\hat{V}}_{1:T}$ by the VQ-VAE decoder ($D_v$), which also has a symmetric structure with $E_v$. 
The entire model is trained with the following loss:
\begin{multline}\label{eq:vqvaeloss}
\calL_{\text{VQ}}=\norm{\mathbf{V}_{1:T}-\hat{\mathbf{V}}_{1:T}}_1\\
+\norm{\texttt{sg}(\mathbf{\hat{Z}})-\mathbf{Z^q}}_2^2+\lambda\norm{\mathbf{\hat{Z}}-\texttt{sg}(\mathbf{Z^q})}_2^2,
\end{multline}
where the first term is the motion reconstruction loss, the latter two terms are for updating the codebook, \texttt{sg} is a stop gradient operation~\cite{sg}, and $\lambda$ is a weight factor.
After training the codebook of discrete facial motions, these discrete motions are utilized in the subsequent stage to learn the speech-conditioned synthesis of 3D facial movements.

\paragraph{Learning speech-driven motion synthesis}
In this stage, we develop a model that maps input speech to a sequence of discrete codes, which are later decoded into realistic continuous motions (refer to Stage 2 in \Fref{fig:pipeline}).
As we target to handle multiple languages, we adopt a multilingual speech encoder $E_m$, pretrained on 53 languages~\cite{wav2vecm}. 
This enables the model to extract language-agnostic speech representations from the multilingual inputs.
Moreover, the model incorporates a language style embedding $\boldsymbol{l}$ alongside the speech. 
Each language style embedding is learnable, effectively capturing the distinct facial movement style associated with speaking in that particular language.

Conditioned on both the input speech and the language style embedding, a Transformer decoder $D_m$ is trained to autoregressively generate the sequence of discrete facial motions.
The Transformer decoder is equipped with the causal self-attention that learns the dependencies within the sequence of previous facial motions and uses cross-modal attention to align the audio with the facial motions. 
The autoregressive modeling process of the Transformer decoder is written as: 
$\mathbf{\hat{z}}_t=D_m(E_m(\mathbf{S}_{1:T}),\boldsymbol{l}, \mathbf{\hat{V}}_{1:t-1})$, where $\mathbf{\hat{z}}_t$ is the currently predicted discrete facial motion, and $\mathbf{\hat{V}}_{1:t-1}$ is the past predicted sequences. 
The discrete code $\mathbf{\hat{z}}_t$ is then quantized by \Eref{eq:quantize} and decoded to continuous motion, $\mathbf{\hat{v}}_t=D_v(Q_v(\mathbf{\hat{z}}_t))$.
The model is trained in a teacher-forcing manner with the following loss:
\begin{equation}\label{eq:stage2}
\calL_{\text{GEN}}=\norm{\mathbf{\hat{Z}}_{1:T}-\texttt{sg}(\mathbf{Z^q}_{1:T})}^2_2+\norm{\mathbf{\hat{V}}_{1:T}-\mathbf{V}_{1:T}}^2_2,
\end{equation}
where the first term regularizes the deviation between the predicted motion features $\mathbf{\hat{Z}}_{1:T}$ and the quantized features $\mathbf{Z^q}_{1:T}$ from the codebook, while the latter term denotes the reconstruction loss between the predicted facial motions $\mathbf{\hat{V}}_{1:T}$ and the ground-truth $\mathbf{V}_{1:T}$.

\paragraph{Implementation details}
The VQ-VAE in the first stage is trained for 150 epochs with the AdamW optimizer ($\beta_1=0.9$, $\beta_2=0.999$ and $\epsilon=10^{-8}$), where the learning rate is initialized as $10^{-4}$, and a mini-batch size of 1.
In the second stage, the Transformer decoder model is trained for 100 epochs with the Adam optimizer, maintaining the same hyper-parameters as in the first stage.
Training for both stages is conducted on a single GeForce RTX 3090 GPU.
\section{Experiments}
In the experiment, we aim to demonstrate the efficacy of our proposed dataset and the baseline model, MultiTalk, in enhancing the multilingual capabilities of 3D talking heads. 
To this end, we compare MultiTalk trained on our dataset, against competing models~\cite{voca, faceformer, codetalker, selftalk} trained on the existing dataset. Specifically, MultiTalk is trained on a subset of our proposed dataset, comprising approximately 20 hours of 3D facial sequences in diverse languages. In contrast, existing works utilize the VOCASET dataset~\cite{voca}, which includes 480 sequences (approximately 30 minutes) only in English from 12 subjects.
We construct a test split in our MultiTalk dataset, involving 60 clips across 12 different languages. To ensure a fair comparison, all results are standardized to the same FLAME topology.

\begin{table}[tp]
  \caption{\textbf{Preliminary experiment.} Audio-Visual Lip Readability (AVLR) demonstrates a high correlation with human evaluations, indicating its suitability as a metric for measuring the lip readability of 3D talking heads. The top three rows present WER (\%) obtained from AVLR and VSR. The last row shows Spearman's correlation coefficient, which ranges from -1 to 1; a value of 1 indicates the highest correlation with human evaluations.}
  \vspace{-2mm}
\label{tab:eval}
\centering
    \resizebox{1\linewidth}{!}{
        \begin{tabular}{l@{\quad\,\,\,}c@{\quad\,\,\,}c@{\quad\,\,\,}c@{\quad\,\,\,}c}
        \toprule
        Method & AVLR (SNR=-7.5) & AVLR (SNR=-10)& VSR\\
        \cmidrule{1-5}
        VOCASET (GT) & 39.4& 43.8& 111.2&\\
        FaceFormer & 50.7& 56.9& 136.3&\\
        VOCA & 53.1 & 62.6 & 153.1 &\\
        \cmidrule{1-5}
        Spearman's $\rho$ & \textbf{0.55}& 0.46& 0.43&\\
        \bottomrule
        \end{tabular}
        }
        \vspace{-2mm}
\end{table}
\subsection{Comparison with existing methods}
To evaluate lip synchronization of generated mesh with the input speech, we measure the Lip Vertex Error (LVE) metric proposed in MeshTalk~\cite{meshtalk}. However, solely measuring the $\ell_2$ error of lip vertices is insufficient for assessing the facial movement due to the one-to-many mapping nature of this task. As a complementary, we introduce a new metric to evaluate the lip readability of the generated mesh, \ie, audio-visual lip readability.

\paragraph{Lip Vertex Error (LVE)} LVE computes the average $\ell_2$ error between the lip regions of the generated mesh vertices and the ground-truth from the test set. For each frame, the LVE is defined as the maximum $\ell_2$ error across all lip vertices.

\paragraph{Audio-Visual Lip Readability (AVLR)} 
We propose the AVLR metric for evaluating perceptually accurate lip readability with a pre-trained Audio-Visual Speech Recognition (AVSR)~\cite{muavic} model.
A na\"ive way for assessing lip readability would be to use a pre-trained Visual Speech Recognition (VSR) model to measure the lip reading metric on the rendered 3D faces without accompanying speech. 
However, relying solely on visual cues introduces ambiguity in inferring words. For example, distinguishing between ``ba''and ``ma'' is challenging by merely observing mouth shapes~\cite{mcgurk1976hearing}. We hypothesize that supplementing visual information with subtle audio cues may reduce this ambiguity, leading to a more robust lip readability metric compared to using visual cues alone.
Specifically, we supply noisy audio alongside the rendered 3D faces to a pre-trained AVSR model and measure the Word Error Rate (WER) to evaluate the lip readability of the 3D talking head.

To validate our proposed AVLR metric, we conduct a preliminary experiment. We collect meshes from the ground-truth VOCASET~\cite{voca} dataset and those generated by FaceFormer~\cite{faceformer} and VOCA~\cite{voca}. 
We measure the WER for each model using VSR, and AVSR with Signal-to-Noise Ratio (SNR) settings of -7.5 and -10, and subsequently rank the models by their WERs. 
We then compute the Spearman's correlation coefficient, $\rho$, to compare the model rankings with human evaluation rankings. 
As shown in Table \ref{tab:eval}, AVSR exhibits the highest correlation with human evaluations. Furthermore, VSR produces WERs exceeding 110\%, confirming its unsuitability as a metric. These findings highlight the efficacy of our proposed AVLR metric in assessing lip accuracy.

Utilizing the metrics described above, we conduct a quantitative comparison of our MultiTalk model against four different approaches: VOCA~\cite{voca}, FaceFormer~\cite{faceformer}, Codetalker~\cite{codetalker}, and SelfTalk~\cite{selftalk}. 
\Tref{tab:quan} summarizes the LVE and AVLR over the test set of the MultiTalk dataset, with AVLR assessed across four different languages. 
Notably, MultiTalk achieves superior performance compared to the other methods across all metrics. 
Specifically, in the AVLR, the recent method SelfTalk shows comparable performance in English, but MultiTalk excels in languages other than English. 
These results highlight the effectiveness of both our proposed method and the dataset in establishing multilingual capabilities for 3D talking head models.

\begin{figure}[t]
    \centering
    \includegraphics[width=1\linewidth]{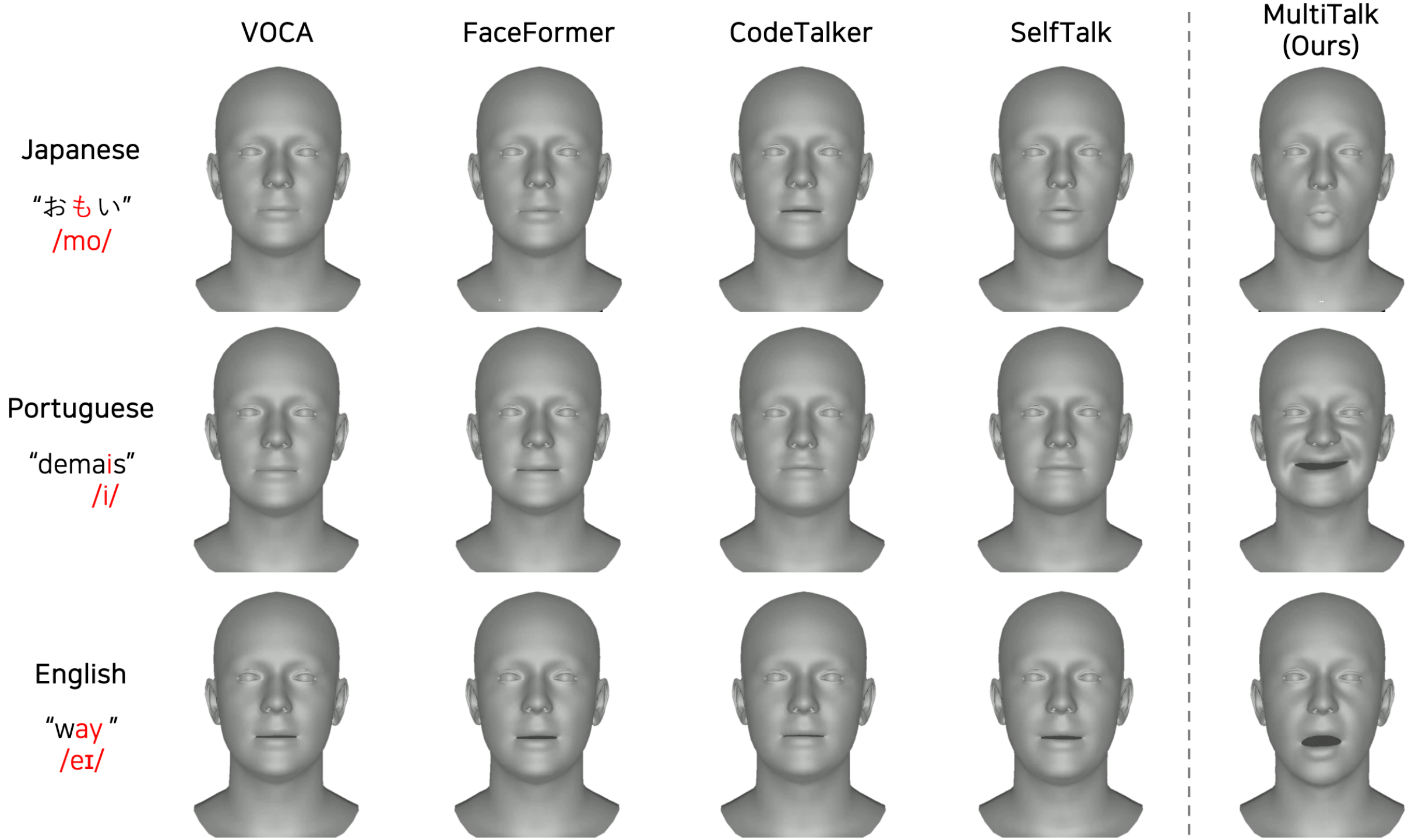}
    \vspace{-4mm}
    \caption{\textbf{Qualitative comparisons.} Compared to existing methods, MultiTalk (Ours) demonstrates detailed facial expressions with accurately synchronized lip movements to the input speech.}\vspace{-3mm}
    \label{fig:qual}
\end{figure}

For a more comprehensive comparison, we visualize the generated samples in \Fref{fig:qual}. 
As shown, the meshes generated by our MultiTalk model exhibit detailed and expressive lip movements for closures and openings in sync with the input speech.
We postulate that such expressiveness could be learned from our dataset, which reflects the inherent diversity and includes various facial movements across languages.

\begin{table}[tp]
  \caption{\textbf{Quantitative comparison to existing methods.} We compare MultiTalk (Ours) with existing methods on the test split of the MuliTalk dataset on 4 languages: English (En), Italian (It), French (Fr), and Greek
(El). LVE is measured in $\times 10^{-4}\text{mm}$ scale, and AVSR (SNR=-7.5) is measured in WER (\%).
  }\vspace{-2mm}
  \label{tab:quan}
  \centering
    \resizebox{1\linewidth}{!}{
        \begin{tabular}{l@{\quad\,\,}c@{\quad\,\,}c@{\quad\,\,}c@{\quad\,\,}c@{\quad\quad}c@{\quad}c@{\quad\,\,}c@{\quad\,\,}c}
        \toprule
        \multirow{2}{*}{Method}&\multicolumn{4}{c}{LVE ($\downarrow$)}&\multicolumn{4}{c}{AVLR ($\downarrow$)}\\
        & En & It & Fr & El &En & It  & Fr & El\\
        \cmidrule{1-9}
        
        VOCA & 1.95& 2.78&1.93& 2.18& 50.8&60.4&74.9&82.1\\
        FaceFormer & 1.82& 2.56&1.78& 1.99& 50.8&58.9&70.9&79.0\\
        CodeTalker & 1.98& 2.56&1.99& 2.09& 50.0&59.4&74.9&77.8\\
        SelfTalk & 1.99& 2.59&1.98& 2.11& 42.8&56.5&68.3&80.3\\
        MultiTalk (Ours) & \textbf{1.16}& \textbf{1.06}&\textbf{1.39}& \textbf{1.26}& \textbf{42.4}&\textbf{50.5}&\textbf{63.0}&\textbf{74.2}\\
        \bottomrule
        \end{tabular}
        }
\end{table}

\begin{table}[tp]
  \caption{\textbf{User study results.} We adopt A vs. B test and report the percentage (\%) of preferences for A (Ours) over B, assessing the generated meshes on lip sync and realism.}\vspace{-2mm}
\label{tab:human}
\centering
    \resizebox{1\linewidth}{!}{
        \begin{tabular}{l@{\quad\,\,}c@{\quad\,\,}c@{\quad\,\,}c@{\quad\,\,}c}
        \toprule
        Aspect & \textit{vs.} VOCA & \textit{vs.} FaceFormer & \textit{vs.} CodeTalker & \textit{vs.} SelfTalk \\
        \cmidrule{1-5}
        Lip sync. & 93.28& 84.17& 76.92& 53.78\\
        Realism. & 93.28& 91.53& 82.05& 66.95\\
        \bottomrule
        \end{tabular}
        }
\end{table}

\begin{table}[tp]  
  \caption{\textbf{Ablation studies on design choices.} We compare different configurations of our method by either incorporating the language style embedding $\boldsymbol{l}$ or utilizing different speech encoders $E_m$. ``Multi.'' and ``En.'' denote the speech encoders trained on multilingual and English-only speeches, respectively.}\vspace{-2mm}
  \label{tab:ablation}
  \centering
    \resizebox{1\linewidth}{!}{
        \begin{tabular}{l@{\quad}c@{\quad\,\,}c@{\quad\,\,}c@{\quad\,\,}c@{\quad\,\,}c@{\quad\,\,}c@{\quad\quad}c@{\quad}c@{\quad\,\,}c@{\quad\,\,}c}
        \toprule
        &\multirow{2}{*}{\begin{tabular}[c]{@{}c@{}}$\boldsymbol{l}$\\ emb.\end{tabular}}&\multirow{2}{*}{\begin{tabular}[c]{@{}c@{}}$E_m$\\ type\end{tabular}}&\multicolumn{4}{c}{LVE ($\downarrow$)}&\multicolumn{4}{c}{AVLR ($\downarrow$)}\\
        &&&  En & It & Fr & El &En & It  & Fr & El\\
        \cmidrule{1-11}
        (a) & &Multi. & 1.78& 2.07&2.45&1.82& \textbf{41.9}& 52.4&64.1&\textbf{72.0}\\
        (b) &\checkmark &En. & 1.56& 1.34&1.91&1.37& 50.3& 55.6&71.7&77.3\\
        (c) &\checkmark &Multi. &\textbf{1.16}& \textbf{1.06}&\textbf{1.39}& \textbf{1.26}& 42.4&\textbf{50.5}&\textbf{63.0}&74.2\\
        \bottomrule
        \end{tabular}
        }\vspace{-3mm}
\end{table}

\subsection{User study}
We incorporate human perception as a metric through a user study. 
We first generate 24 3D face videos using MultiTalk (A) and compare them with those generated by existing methods (B) from the test split of the MultiTalk dataset. 
We design an A vs. B test, prompting participants to choose between two samples based on lip synchronization and realism. 
To accurately evaluate multilingual capability, participants from various countries participated in this study. 
As indicated in \Tref{tab:human}, MultiTalk is preferred by users, notably excelling in realism compared to other methods.
These results emphasize the expressiveness and multilingual capability of our model.

\subsection{Ablation study}
We conduct ablation studies to validate our design choices, as in \Tref{tab:ablation}. Comparing (a) and (c), we observe that utilizing the language style embedding stabilizes the learning and yields favorable performance. Moreover, comparisons between (b) and (c) indicate that incorporating the multilingual speech encoder facilitates the extraction of language-agnostic features. This enables the model to focus on universal speech representations, thereby accommodating motion synthesis from multiple languages.
\section{Discussion and conclusion}
In this work, we introduce a novel task of animating 3D talking heads from multilingual speeches.
Recognizing the lack of diversity in existing datasets for learning multilingual capabilities,
we have collected the MultiTalk dataset, consisting of 2D talking videos in multiple languages, each paired with 3D metadata and transcripts.
Moreover, we present MultiTalk, a baseline model trained in two stages on our dataset.
Considering the novelty of this task, we have devised an audio-visual lip readability metric to assess the model's multilingual capability. 
Our experiments demonstrate the effectiveness of our approach, showcasing robust lip synchronization performance across diverse languages.

\paragraph{Limitation and future work}
While our dataset offers extensive annotations, the transcripts and 3D meshes are pseudo-annotated, which might introduce some level of noise compared to human annotations. Despite this limitation, these pseudo-annotations have proven to be effective in enhancing model performance in prior arts~\cite{yeo2023visual, autoavsr, laughtalk, emote}. Future work will focus on refining these annotations.
We would like to note that our proposed multilingual video dataset and the Audio-Visual Lip Readablity metric have broader usage beyond our immediate task, \eg, (audio) visual speech recognition and 2D talking heads.
Furthermore, the rich facial motion prior learned by diverse faces across various languages holds significant potential to advance research in facial motion synthesis for further exploration.

\section{Acknowledgments}
This research was supported by a grant from KRAFTON AI, and also partially supported by Institute of Information \& communications Technology Planning \& Evaluation (IITP) grant funded by the Korea government (MSIT) (RS-2022-II220124, Development of Artificial Intelligence Technology for Self-Improving Competency-Aware Learning Capabilities; RS-2021-II212068, Artificial Intelligence Innovation Hub; RS-2019-II191906, Artificial Intelligence Graduate School Program (POSTECH)).

\bibliographystyle{IEEEtran}
\bibliography{mybib}

\end{document}